\documentclass[conference]{IEEEtran}
\IEEEoverridecommandlockouts
\usepackage{cite}
\usepackage{multirow}
\usepackage{amsmath,amssymb,amsfonts}
\usepackage{graphicx}
\usepackage{bbm}
\usepackage{textcomp}
\usepackage{xcolor}
\def\BibTeX{{\rm B\kern-.05em{\sc i\kern-.025em b}\kern-.08em
    T\kern-.1667em\lower.7ex\hbox{E}\kern-.125emX}}

\usepackage[noend]{algpseudocode}
\usepackage{algorithm, algorithmicx}
\usepackage{url}
\begin{document}

\title{Swapped Face Detection using Deep Learning and Subjective Assessment\\
}

% \author{\IEEEauthorblockN{1\textsuperscript{st} Xinyi Ding}
% \IEEEauthorblockA{\textit{Department of Computer Science} \\
% \textit{Southern Methodist University}\\
% Dallas, USA\\
% xding@smu.edu}
% \and
% \IEEEauthorblockN{2\textsuperscript{nd} Zohreh Raziei}
% \IEEEauthorblockA{\textit{Department of Engineering Management, Information and Systems} \\
% \textit{Southern Methodist University}\\
% Dallas, USA \\
% zraziei@mail.smu.edu}
% \and
% \IEEEauthorblockN{3\textsuperscript{rd} Eric C. Larson}
% \IEEEauthorblockA{\textit{Department of Computer Science} \\
% \textit{Southern Methodist University}\\
% Dallas, USA \\
% eclarson@lyle.smu.edu}
% \and
% \IEEEauthorblockN{4\textsuperscript{th} Eli V Olinick}
% \IEEEauthorblockA{\textit{Department of Engineering Management, Information and Systems} \\
% \textit{Southern Methodist University}\\
% Dallas, USA \\
% olinick@lyle.smu.edu}
% \and
% \IEEEauthorblockN{5\textsuperscript{th} Paul Krueger}
% \IEEEauthorblockA{\textit{Department of Mechanical Engineering} \\
% \textit{Southern Methodist University}\\
% Dallas, USA \\
% pkrueger@lyle.smu.edu}
% \and
% \IEEEauthorblockN{6\textsuperscript{th} Michael Hahsler}
% \IEEEauthorblockA{\textit{Department of Engineering Management, Information and Systems} \\
% \textit{Southern Methodist University}\\
% Dallas, USA \\
% mhahsler@lyle.smu.edu}
% }

\author{
    \IEEEauthorblockN{Xinyi Ding\IEEEauthorrefmark{1}, Zohreh Raziei\IEEEauthorrefmark{2}, Eric C. Larson\IEEEauthorrefmark{1}, Eli V. Olinick\IEEEauthorrefmark{2}, Paul Krueger \IEEEauthorrefmark{3}, Michael Hahsler\IEEEauthorrefmark{2}} 
    
    \IEEEauthorblockA{\IEEEauthorrefmark{1}Department of Computer Science,
    Southern Methodist University
    \\ xding@mail.smu.edu, eclarson@lyle.smu.edu}
    \IEEEauthorblockA{\IEEEauthorrefmark{2}Department of Engineering Management, Information and Systems, Southern Methodist University
    \\ zraziei@mail.smu.edu, \{olinick, mhahsler\}@lyle.smu.edu}
     \IEEEauthorblockA{\IEEEauthorrefmark{3}Department of Mechanical Engineering,
    Southern Methodist University
    \\ pkrueger@lyle.smu.edu}
}
\maketitle

\begin{abstract}

The tremendous success of deep learning for imaging applications has resulted in numerous beneficial advances. Unfortunately, this success has also been a catalyst for malicious uses such as photo-realistic face swapping of parties without consent. Transferring one person's face from a source image to a target image of another person, while keeping the  image photo-realistic overall has become increasingly easy and automatic, even for individuals without much knowledge of image processing. In this study, we use deep transfer learning for face swapping detection, showing true positive rates $>$96\% with very few false alarms. Distinguished from existing methods that only provide detection accuracy, we also provide uncertainty for each prediction, which is critical for trust in the deployment of such detection systems. Moreover, we provide a comparison to human subjects. To capture human recognition performance, we build a website to collect pairwise comparisons of images from human subjects. Based on these comparisons, images are ranked from most real to most fake. We compare this ranking to the outputs from our automatic model, showing good, but imperfect, correspondence with linear correlations $>0.75$. Overall, the results show the effectiveness of our method. As part of this study, we create a novel, publicly available dataset that is, to the best of our knowledge, the largest public swapped face dataset created using still images.  Our goal of this study is to inspire more research in the field of image forensics through the creation of a public dataset and initial analysis.
\end{abstract}


\section{Introduction}

Face swapping refers to the process of transferring one person's face from a source image to another person in a target image, while maintaining photo-realism. It has a number of applications in cinematic entertainment and gaming. However, in the wrong hands, this method could also be used for fraudulent or malicious purposes. For example, ``DeepFakes'' is such a project that uses generative adversarial networks (GANs) \cite{goodfellow2014generative} to produce videos in which people are saying or performing actions that never occurred. While some uses without consent might seem benign such as placing Nicolas Cage in classic movie scenes, many sinister purposes have already occurred. For example, a  malicious use of this technology involved a number of attackers creating pornographic or otherwise sexually compromising videos of celebrities using face swapping \cite{Deepfake15:online}. A detection system could have prevented this type of harassment before it caused any public harm.

Conventional ways of conducting face swapping usually involve several steps. %\cite{bitouk2008face}. %More appropriate cites here?
A face detector is first applied to narrow down the facial region of interest (ROI). Then, the head position and facial landmarks are used to build a perspective model. To fit the source image into the target ROI, some adjustments need to be taken. Typically these adjustments are specific to a given algorithm. Finally, a blending happens that fuses the source face into the target area. This process has historically involved a number of mature techniques and careful design, especially if the source and target faces have dramatically different position and angles (the resulting image may not have a natural look). 
% ERIC COMMENT: Whay bring up Bitouk here in the introduction? Seems ouf of place, as such I am commenting this reference 
%In \cite{bitouk2008face}, Bitouk \textit{et al.} searched a database to find a ``best-matching'' image with similar head pose and lighting. Therefore, their method does not allow swapping arbitrary faces.  To swap any two faces, other digital tools are needed and often are conducted manually. %This kind of searching is limited, because we can not freely swap any two images. 

The impressive progress deep learning has made in recent years is changing how face swapping techniques are applied from at least two perspectives. Firstly, models like convolutional neural networks allow more accurate face landmarks detection, segmentation, and pose estimation. Secondly, generative models like GANs \cite{goodfellow2014generative} combined with other techniques like Auto-Encoding~\cite{schmidhuber2015deep}
allow automation of facial expression transformation and blending, making large-scale, automated face swapping possible. Individuals that use these techniques require little training to achieve photo-realistic results. In this study, we use two methods to generate swapped faces~\cite{nirkin2018face, GitHubsh94:online}. Both methods exploit the advantages of deep learning methods using contrasting approaches, discussed further in the next section. We use this dataset of swapped faces to evaluate and inform the design of a face-swap detection classifier.

% ERIC COMMENT: What is the purpose of this paragraph here? I am having trouble parsing the comparison and differences pointed toward other works.

% XINYI COMMENT: Made some changes, the main goal is to introduce some existing swapped faces detection techniques and their limitations

% ERIC COMMENT: Great! It works much better now.

\begin{figure*}[t!]
\centering
\centerline{\includegraphics[width=0.9\linewidth]{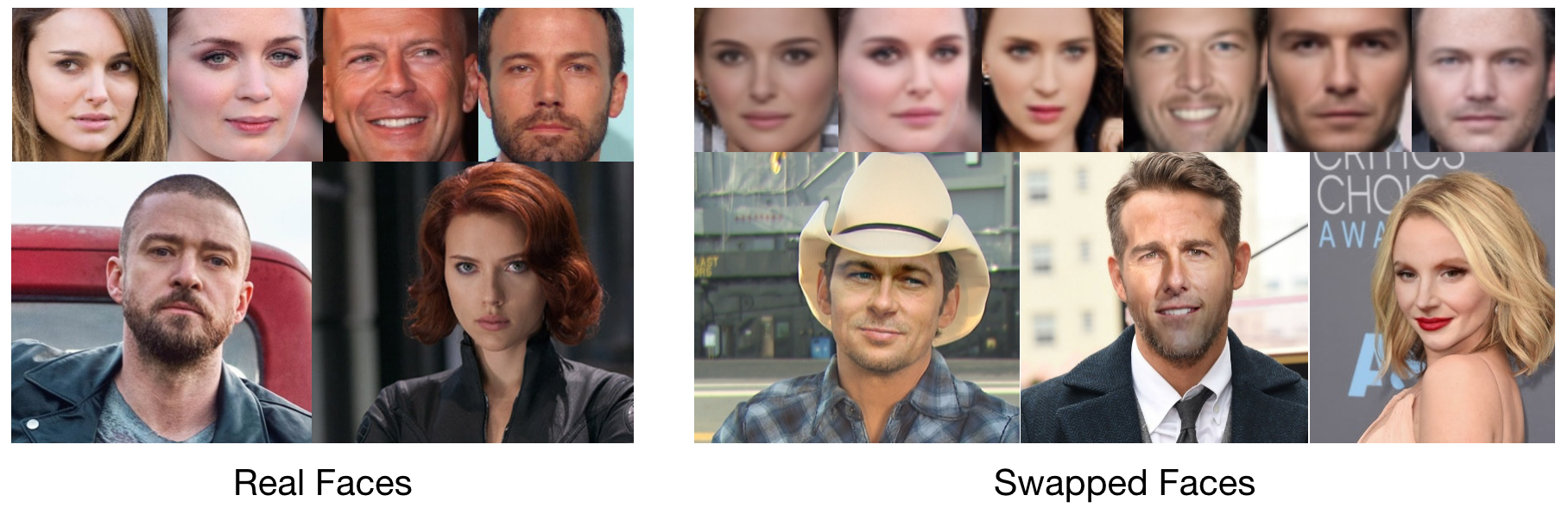}}
\caption{Real and swapped faces in our dataset.  \textbf{Top Row Right}: Auto-Encoder-Gan.  \textbf{Bottom Row Right}: Nirkin's Method }
\label{fig::dataset}
\end{figure*}

With enough data, deep learning based classifiers can typically achieve low bias due to their ability to represent complex data transformations. However, in many cases, the confidence levels of these predictions are also important, especially when critical decisions need to be made based on these predictions. The uncertainty of a prediction could indicate when other methods could be more reliable. Bayesian Deep Learning, for example, assumes a prior distribution of its parameters $P(\mathbf{w})$ and integrates the posterior distribution $P(\mathbf{w}|\mathcal{D})$ when making a prediction, given the dataset $\mathcal{D}$. However, it is usually intractable for models like neural networks and must be employed using approximations to judge uncertainty. We propose a much simpler approach by using the raw logits difference of the neural network outputs (\textit{i.e.}, the odds ratio). We assume, in a binary classification task, if the model has low confidence about a prediction the difference of the two logits output should be small compared with high confidence prediction. We also show that the odds ratio of the neural network outputs is correlated with the human perception of ``fake'' versus ``real.''

The end goal of malicious face swapping is to fool a human observer. Therefore, it is important to understand how human subjects perform in recognizing swapped faces. To this end, we not only provide the accuracy of human subjects in detecting fake faces, but we also provide the ranking of these images from most real to most fake using pairwise comparisons. We selected 400 images and designed a custom website to collect human pairwise comparisons of images. Approximate ranking is used \cite{heckel2018approximate} to help reduce the number of needed pairwise comparisons. With this ranking, we compare the odds ratio of our model outputs to the ranking from human subjects, showing good, but not perfect correspondence. We believe future works can improve on this ranking comparison, providing a means to evaluate face swapping detection techniques that more realistically follow human intuition.  

We enumerate our contributions as follows:

\begin{itemize}
    \item A public dataset comprising 86 celebrities using 420,053 images. This dataset is created using still images, different from other datasets created using video frames that may contain highly correlated images. In this dataset, each celebrity has approximately 1,000 original images (more than any other celebrity dataset). We believe our dataset is not only useful for swapped face detection, it may also be beneficial for developing facial models. 
    
    \item We investigate the performance of two representative face swapping techniques and discuss limitations of each approach. For each technique, we create thousands of swapped faces for a number of celebrity images.
    
    \item We build a deep learning model using transfer learning for detected swapped faces. To our best knowledge, it is the first model that provides high accuracy predictions coupled with an analysis of uncertainties. 
    
    \item We build a website that collects pairwise comparisons from human subjects in order to rank images from most real to most fake. Based on these comparisons, we approximately rank these images and compare to our model. 
\end{itemize}

\section{Related Work}

There are numerous existing works that target face manipulation and detection. Strictly speaking, face swapping is simply one particular kind of image tampering. Detection techniques designed for general image tampering may or may not work on swapped faces, but we expect specially designed techniques to perform superior to generic methods. Thus, we only discuss related works that directly target or involve face swapping and its detection. 

\subsection{Face Swapping}
Blanz \textit{et al.} \cite{blanz2004exchanging} use an algorithm that estimates a 3D textured model of a face from one image, applying a new facial ``texture'' to the estimated 3D model. The estimations also include relevant parameters of the scene, such as the orientation in 3D, camera's focal length, position, illumination intensity, and direction. The algorithm resembles the Morphable Model such that it optimizes all parameters in the model in conversion from 3D to image. 
Bitouk \textit{et al.} \cite{bitouk2008face}, bring the idea of face replacement without the use of 3D reconstruction techniques. The approach involves the finding of a candidate replacement face which has similar appearance attributes to an input face. It is, therefore, necessary to create a large library of images. A ranking algorithm is then used in selecting the image to be replaced from the library. To make the swapped face more realistic, lighting and color properties of the candidate images might be adjusted. Their system is able to create subjectively realistic swapped faces. However, one of the biggest limitations is that it is unable to swap an arbitrary pair of faces. 
Mahajan \textit{et al.} \cite{mahajan2017swapitup} present an algorithm that automatically chooses faces that are facing the front and then replaces them with stock faces in a similar fashion, as Bitouk \textit{et al.} \cite{bitouk2008face}. 

% The empirical success of deep learning in image processing has also resulted in a bunch of new face swapping techniques is that the Mahajan's algorithm does not make estimations of the illumination, nor does it make corrections.

Chen \textit{et al.} \cite{chen2019face} suggested an algorithm that can be used in the replacement of faces in referenced images that have common features and shape as the input face. A triangulation-based algorithm is used in warping the image by adjusting the reference face and its accompanying background to the input face. A parsing algorithm is used in accurate detection of face-ROIs and then the Poisson image editing algorithm is finally used in the realization of boundaries and colour correction. 
Poisson editing is explored from its basics by Perez \textit{et al.} \cite{perez2003poisson}. Once given methods to craft a Laplacian over some domain for an unknown function, a numerical solution of the Poisson equation for seamless domain filling is calculated. This technique can independently be replicated in color image channels.

The empirical success of deep learning in image processing has also resulted in many new face swapping techniques. Korshunova \textit{et al.}\cite{korshunova2017fast} approached face swapping as a style transfer task. They consider pose and facial expression as the content and identity as the style. A convolutional neural network with multi-scale branches working on different resolutions of the image is used for transformation. Before and after the transformation, face alignment is conducted using the facial keypoints. 
Nirkin \textit{et al.} \cite{nirkin2018face} proposed a system that allows face swapping in more challenging conditions (two faces may have very different pose and angle). They applied a multitude of techniques to capture facial landmarks for both the source image and the target image, building 3D face models that allow swapping to occur via transformations. A fully convolutional neural network (FCN) is used for segmentation and for blending technique after transformation. 

% PAUL COMMENT:  The first sentence in this paragraph is a little disjointed.  Maybe it should read "The popular approaches of Auto-Encoders and generative adversarial networks (GANs) can make face swapping more automated..."  Also, in the next to the last sentence in this paragraph, CNN should be defined.

The popularity of Auto-Encoder\cite{schmidhuber2015deep} and generative adversarial networks (GANs) \cite{goodfellow2014generative} makes face swapping more automated, requiring less expert supervision. A variant of the DeepFake project is based on these two techniques \cite{GitHubsh94:online}. The input and output of an Auto-Encoder is fixed and a joint latent space is discovered. During training, one uses this latent space to recover the original image of two (or more) individuals. Two different auto-encoders are trained on two different people, sharing the same Encoder so that the latent space is learned jointly. This training incentivizes the encoder to capture some common properties of the faces (such as pose and relative expression). The decoders, on the other hand, are separate for each individual so that they can learn to generate realistic images of a given person from the latent space. 
Face swapping happens when one encodes person A's face, but then uses person B's decoder to construct a face from the latent space. The variant of this method in \cite{GitHubsh94:online} uses an auto-encoder as a generator and a CNN as the discriminator that checks if the face is real or swapped. Empirical results show that adding this adversarial loss  improves the quality of swapped faces.

%%%%%%%%%%%%%%%%%%%%%%%%%%%%%

%  The algorithm never poses for estimations or corrections \cite{mahajan2017swapitup} and \cite{lin2012face}.}

% \textcolor{red}{Once given methods to craft a Laplacian over some domain for unknown function, numerical solution of the Poisson equation for seamless domain filling. This technique can independently be replicated in colour image channels. This process paves the way for swapping of face \cite{perez2003poisson}. Blanz \textit{et al.} \cite{blanz2004exchanging}, uses an algorithm that estimates a 3D textured model of a face from one image. The estimations also include relevant parameters of the scene, such as the orientation in 3D, camera's focal length, position and illumination intensity and direction. The algorithm resembles the Morphable Model such that it optimizes all parameters in the model in conversion from 3D to image.}

% \textcolor{red}{Bitouk \textit{et al.} \cite{bitouk2008face}, bring the idea of face replacement without the use of 3D reconstruction techniques. The approach involves the finding of a candidate replacement face which has similar appearance attributes to an input face. It is, therefore, necessary to create a large library of images. A ranking algorithm is then used in selecting the image to be replaced from the library. 

%  The algorithm never poses for estimations or corrections \cite{lin2012face} and \cite{mahajan2017swapitup}.

Natsume \textit{et al.} \cite{natsume2018rsgan} suggests an approach that uses hair and faces in the swapping and replacement of faces in the latent space. The approach applies a generative neural network referred to as an RS-GAN (Region-separative generative adversarial network) in the generation of a single face-swapped image. 
Dale \textit{et al.} \cite{dale2011video} bring in the concept of face replacement in a video setting rather than in an image. In their work, they use a simple acquisition process in the replacement of faces in a video using inexpensive hardware and less human intervention. 

% This work is about object contor detection
% Yang \textit{et al.} \cite{yang2016object} further the use of convolution in the detection of the image contour. The immediate application of image contour detection is the image or face swapping. A convolution encoder-decoder network (CEDN) can be developed thanks to the success of the fully deconvolution and convolution networks in semantic segmentation. A VGG-16 net initializes the encoder, and the decoder is constructed with alternated unpooled convolution layers for dense prediction of the image size.

%% This work is about facial features localization,
%% 
% Interactive facial feature localization is another interesting image swapping feature that is discussed by Le \textit{et al.} \cite{le2012interactive}. An algorithm is designed and used in the accommodation of partial labels in the process of localization of novel facial components. An extension of the algorithm is provided to take care of incomplete observations in the localization process. In finalizing the process, a challenging dataset of localization of facial components is introduced at a high resolution of 2330 \cite{le2012interactive}.}

\subsection{Fake Face Detection}

Zhang \textit{et al.} \cite{zhang2017automated} created swapped faces using labeled faces in the wild (LFW) dataset \cite{LFWTech}. They used sped up robust features (SURF) \cite{bay2006surf} and Bag of Words (BoW) to create image features instead of using raw pixels. After that, they tested on different machine learning models like Random Forests, SVM's, and simple neural networks. They were able to achieve accuracy over 92\%, but did not investigate beyond their proprietary swapping techniques. The quality of their swapped faces is not compared to other datasets. Moreover, their dataset only has 10,000 images (half swapped) which is relatively small compared to other work.  

%PAUL COMMENT:  I'm not sure if the sentence "They evaluated several detection methods of DeepFakes." is correct.  Should it be "They evaluated several detection methods for DeepFakes"?
Khodabakhsh \textit{et al.} \cite{khodabakhsh2018fake} examined the generalization ability of previously published methods. They collected a new dataset containing 53,000 images from 150 videos. The swapped faces in their data set were generated using different techniques. Both texture-based and CNN-based fake face detection were evaluated.  Smoothing and blending were used to make the swapped face more photo-realistic. However, the use of video frames increases the similarity of images, therefore decreasing the variety of images. 
Agarwal \textit{et al.} \cite{agarwal2017swapped} proposed a feature encoding method termed as Weighted Local Magnitude Patterns. They targeted videos instead of still images. They also created their own data set. Korshunov \textit{et al.} also targeted swapped faces detection in video \cite{korshunov2018deepfakes}. They evaluated several detection methods of DeepFakes. What's more, they analyze  the vulnerability of VGG and FaceNet based face recognition systems. 

A recent work from R{\"o}ssler \textit{et al.} \cite{rossler2019faceforensics++} provides an evaluation of various detectors in different scenarios. They also report human performance on these manipulated images as a baseline.  Our work shares many similarities with these works. The main difference is that we provide a large scale data set created using still images instead of videos, avoiding image similarity issues. Moreover, we provide around 1,000 different images in the wild for each celebrity. It is useful for models like auto-encoders that require numerous images for proper training. In this aspect, our data set could be used beyond fake face detection. The second difference is that we are not only providing accuracy from human subjects, but also providing the rankings of images from most real to most fake. We compare this ranking to the odds ratio ranking of our classifier showing that human certainty and classifier certainty are relatively (but not identically) correlated.

\section{Experiment}

% \subsection{Set up}
% The great success deep learning has achieved in image processing has been a catalyst for more creative research like face swapping. In this paper, we use two face swapping techniques which are two representatives of currently used: One is a combination of more traditional techniques that involves 3D model building, Fitting, segmentation, transfer etc, which we call PipeLine method [cite]. The other is using generative adversarial network [cite]. This model takes images as input and automatically generate swapped face images. It can be trained end to end using backpropgation. 

\subsection{Dataset}

\begin{table}[]
\centering
\caption{Dataset Statistics}
\begin{tabular}{|l|l|l|l|l|}
\hline
          & Nirkin's Method \cite{nirkin2018face} & AE-GAN  \cite{GitHubsh94:online} & \multicolumn{2}{l|}{Total}        \\ \hline
Real Face & 72,502          & 84,428  & \multicolumn{2}{l|}{156,930} \\ \hline
Swapped Face  & 178,695         & 84,428  & \multicolumn{2}{l|}{263,123} \\ \hline
Total     & 251,197         & 168,856 & \multicolumn{2}{l|}{420,053} \\ \hline
\end{tabular}
\label{tab:dataset}
\end{table}

%PAUL COMMENT:  I recommend putting the data set on SMU Scholar.  They can curate it for you.

Face swapping methods based on auto-encoding typically require numerous images from the same identity (usually several hundreds). There was no such dataset that met this requirement when we conducted this study, thus, we decided to create our own. Access to Version 1.0 of this dataset is freely available at the noted link \footnote{https://www.dropbox.com/sh/rq9kcsg3kope235/AABOJGxV6ZsI4-4bmwMGqtgia?dl=0}. The statistics of our dataset are shown in Table \ref{tab:dataset}.

% \begin{figure}[htbp]
% \centerline{\includegraphics[width=0.8\linewidth]{images/dataset.png}}
% \caption{Cropped images in dataset. Top row: for AutoEncoder-GAN (AE-GAN) method. Bottom row: for Nirkin's method}
% \label{fig::dataset}
% \end{figure}

All our celebrity images are downloaded using the Google image API. After downloading these images, we run scripts to remove images without visible faces and remove duplicate images. Then we perform cropping to remove extra backgrounds. Cropping was performed automatically and inspected visually for consistency. We created two types of cropped images as shown in Figure \ref{fig::dataset} Left.  
One method for face swapping we employed involves face detection and lighting detection, allowing the use of images with larger, more varied backgrounds. On the other hand, another method is more sensitive to the background, thus we eliminate as much background as possible. In a real deployment of such a method, a face detection would be run first to obtain a region of interest, then swapping would be performed within the region. In this study, for convenience, we crop the face tightly when a method requires this.
%, but do not place the swapped face back into the original image, which could be carried out easily  %because blurriness artifacts would make the swap obvious. 

The two face swapping techniques we use in this study are representatives of many algorithms in use today. Nirkin's method \cite{nirkin2018face} is a pipeline of several individual techniques. On the other hand, the Auto-Encoder-GAN (AE-GAN) method is completely automatic, using a fully convolutional neural network architecture \cite{GitHubsh94:online}. In selecting individuals to swap, we randomly pair celebrities within the same sex and skin tone. Each celebrity has around 1,000 original images. For Nirkin's method, once a pair of celebrities is chosen, we randomly choose one image from these 1,000 images as the source image and randomly choose one from the other celebrity as the target image. We noticed, for Nirkin's method, when the lighting conditions or head pose of two images differs too dramatically, the resulted swapped face is of low quality. On the other hand, the quality of swapped faces from the AE-GAN method is more consistent. 
% We manually remove these failure cases in our dataset using visual inspection. 

%% May be just discuss it using texts is enough
% \begin{figure}[tb]
% \centerline{\includegraphics[width=0.9\linewidth]{images/failure_cases.png}}
% \caption{Some failure cases when the source image and target image has very different lighting condition and head pose}
% \label{fig::failure_cases}
% \end{figure}

\subsection{Classifier}

% If we view a neural network as a probabilistic model $P(\mathbf{y}|\mathbf{x},\mathbf{w})$. Given training data $\mathcal{D}$ = \{($\mathbf{x}^i$, $\mathbf{y}^i$), i = 1,...,N\}, we could get the parameter theta by maximum likelihood estimate (MLE):

% \begin{equation}
%     \mathbf{w}^{MLE} = \argmax_{\mathbf{w}} \sum_{i=1}^{N} logP(\mathbf{y}^i| \mathbf{x}^i, \mathbf{w})
% \end{equation}
% We are ignoring regularisation here for simplicity. Global optimal is usually not accessible for deep models. Iterative methods like gradient descent could be used to get a local optimal, such local optimal could usually give satisfying results in practice. 

Existing swapped face detection systems based on deep learning only provide an accuracy metric, which is insufficient for a classifier that is used continuously for detection. Providing an uncertainty level for each prediction is important for the deployment of such systems, especially when critical decisions need to be made based on these predictions. 
%One way to provide uncertainty is to take the bayesian approach.

%Bayesian Deep Learning incorporate uncertainty by using posterior distribution of weights  $P(\mathbf{w}|\mathcal{D})$, given the training data $\mathcal{D}$. Thus, the predictive distribution is given by $P(\hat{\mathbf{y}}|\hat{\mathbf{x}})=\E_{P(\mathbf{w}|\mathcal{D})}[P(\hat{\mathbf{y}}|\hat{\mathbf{x}}, \mathbf{w})]$ for a prediction $\hat{\mathbf{y}}$ of a test item $\hat{\mathbf{x}}$. However, integrating over all possible $\mathbf{w}$ is intractable and an approximate approach is used in practice. 
In this study, we use the odds ratio of the binary classification output (i.e., the raw logits difference of each neural network output) as an uncertainty proxy. For binary classification tasks, the final layer of a deep learning model usually outputs two logits (before sending them to a squashing function). The model picks the large logit as the prediction. We assume if the model is less certain about a prediction, the difference of these two logits should be smaller than that of a more certain prediction. We note that this method is extremely simple as compared to other models that explicitly try to model uncertainty of the neural network, such as Bayesian deep learning methods. The odds ratio, on the other hand, does not explicitly account for model uncertainty of the neural network---especially when images are fed into the network that are highly different from images from the training data. Even so, we find that the odds ratio is an effective measure of uncertainty for our dataset, though more explicit uncertainty models are warranted for future research. 

%Uncertainties could be measured using outputs variance. Kendall \textit{et al.} use Monte Carlo and sample unaries through the softmax function \cite{kendall2017uncertainties}

% \begin{equation}
%     \hat{\mathbf{x}}_{i, t} = \mathbf{f}_i^{\mathbf{W}} + \sigma_i^{\mathbf{W}}\epsilon_t, \epsilon_t \sim \mathcal{N}(0, I)
% \end{equation}

% \begin{equation}
%     \mathcal{L}_x = \sum_{i=1}log\frac{1}{T}\sum_t \exp(\hat{x}_{i,t,c} - log\sum_{c^\prime}\exp\hat{x}_{i,t,c^\prime})
% \end{equation}

% We extend their work and using the following loss function. 

Deep learning methods usually take days or weeks to train. Models like ResNet can easily have tens or hundreds of layers. It is believed with more layers, more accurate hierarchical representation could be learned. Transfer leaning allows us to reuse the learned parameters from one task to another similar task, thus avoiding training from scratch, which can save a tremendous amount of resources. In this study, we apply transfer learning using ResNet-18, which is originally trained to perform object recognition on ImageNet \cite{deng2009imagenet}. Since we are performing binary classification in this study, we replace the final layers of ResNet-18 with custom dense layers and then train the model in stages. During the first stage, we constrain the ResNet-18 architecture to be constant while the final layers are trained. After sufficient epochs, we then ``fine tune'' the ResNet-18 architecture, allowing the weights to be trained via back-propagation for a number of epochs.  

\begin{figure}[t]
\centerline{\includegraphics[width=1.0\linewidth]{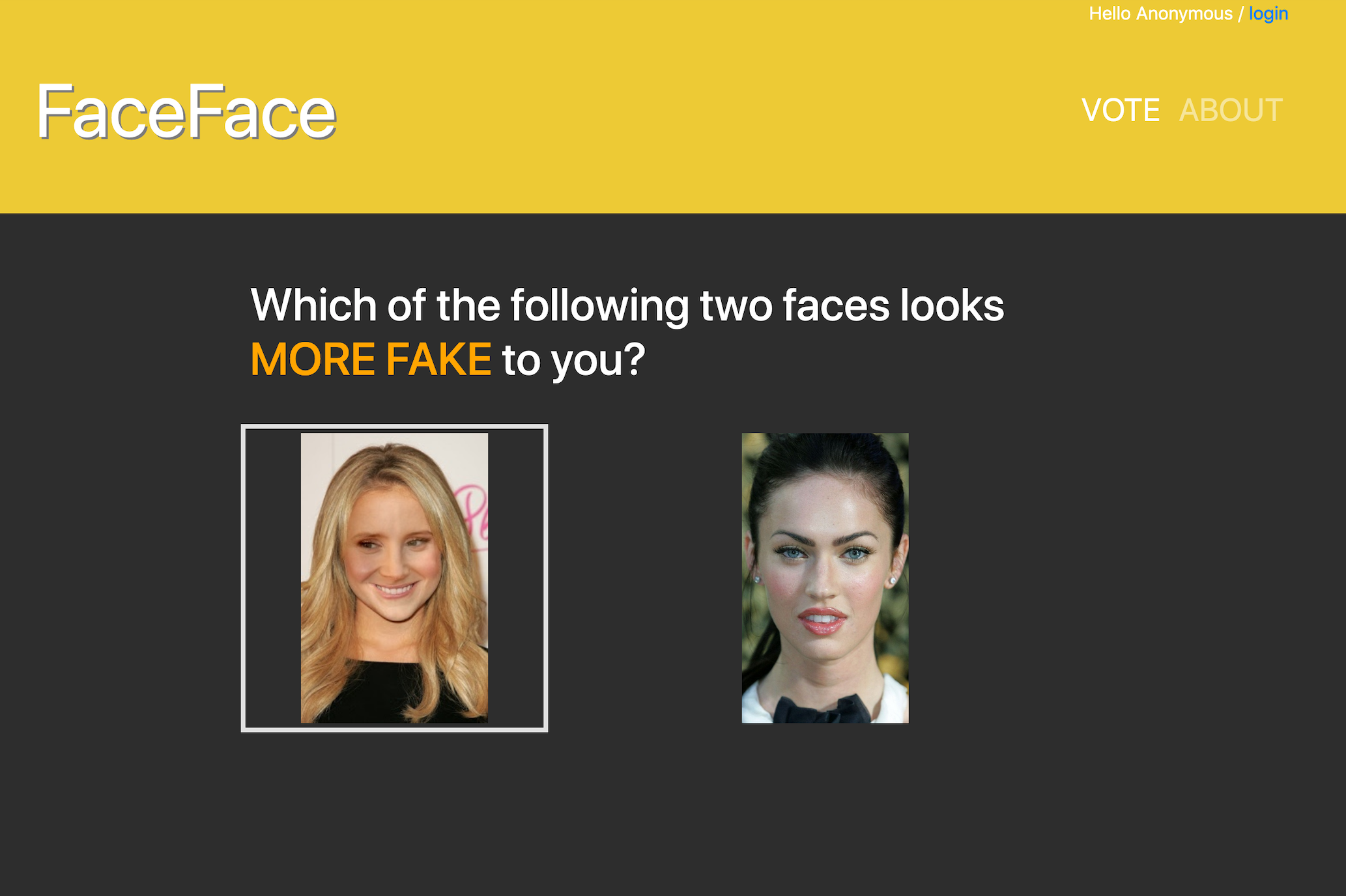}}
\caption{A screenshot of the website collecting comparisons. As the mouse hovers over the left image, it is highlighted}
\label{fig::web}
\end{figure}

\subsection{Human Subjects Comparison}
Because face swapping attacks are typically aimed at misleading observers, it is highly important to understand how human beings perform at detecting swapped faces. Thus, a research contribution should also compare classification with human subjects. In this research it is not only our aim to provide the accuracy of human subjects at detecting swapped faces, but also to establish a ranking of images from most real to most fake. For example, if a rater thinks that an image is fake, is it obvious or is that rater not quite sure about their decision? We argue that this uncertainty is important to model. Moreover, we argue that, if humans are somewhat adept at finding fake images, then the machine learning model should have a similar ranking of the images from most real to most fake. We argue this because the human mind can leverage many information sources and prior knowledge not available to a simple machine learning algorithm. Thus, while two machine learning model may perform  detection perfectly, if one follows human ranking more closely, it can be judged superior. 

However, it is impractical to rank all  image pairs in our dataset with multiple human raters. Therefore we apply two techniques to mitigate the ranking burden. First we only rank a subset of the total images, and, second, we perform approximate ranking of image pairs. As a subset of images, we manually select 100 high-quality swapped faces from each method together with 200 real faces (400 images in total). The manual selection of high quality images is justified because badly swapped faces would be easily recognized. Thus, an attacker would likely perform the same manner of ``re-selecting'' only high quality images before releasing them for a malicious purpose. It is of note that, even with only 400 images, the number of pairwise ratings required for ranking (over 79,000) poses a monumental task.

\subsubsection{Approximate Ranking}
To get the ranking, we designed and deployed a website that implements the approximate ranking algorithm in \cite{heckel2018approximate}. Users on the website are asked to compare two images and select which image appears most fake, subjectively. We  use approximate ranking because even for only 200 images, a full ranking would require more than 19,900 comparisons (with each evaluator ranking every image pair only one time, despite the fact that different subjects may have different opinions). To converge, this method could easily require many more evaluations when two evaluators do not agree (perhaps more than 100,000 pairwise ratings). The approximate ranking algorithm, Hamming-LUCB, helps alleviate this need \cite{heckel2018approximate}.
This algorithm seeks to actively make identification of  two ordered sets of images $S_1$ and $S_2$, representing the highest and lowest ranked images, respectively. For a set $[n]$ of $n$ images, subsets $S_1$ and $S_2$ consist of a range of items of size $k-h$, where $h$  is the ``allowed'' number of mistakes in each set, and the   $n-k-h$  items perceived as least fake comprise the second set. Between the two sets, there is a high confidence that the items contained in the first set score similarly as compared to those items that are contained in the second set. The remaining items, after finding the two sets, can arbitrarily be distributed in high confidence to the two sets such that an accurate (but approximate) Hamming ranking is obtained.
In the algorithm, the two sets are defined on the basis of adaptive definition of estimations of the scores (image rankings) $\tau_i$ for each $i \in [n]$.

% PAUL COMMENT:  need to define $\tau_i$ in paragraph above as the "score" of the image.  Also, the statement in the last sentence above is incomplete.  It should state that a score, $\tau_i$, is obtained for each image.

% \usepackage{multirow}
\begin{table*}[!htbp]
\centering
\caption{Overall Results}
\begin{tabular}{|c|c|c|c|c|c|c|c|}
\hline
\multicolumn{2}{|c|}{\multirow{2}{*}{}}                               & \multicolumn{3}{c|}{Nirkin's Method \cite{nirkin2018face}}                                          & \multicolumn{3}{c|}{AE-GAN \cite{GitHubsh94:online}}                                                   \\ \cline{3-8} 
\multicolumn{2}{|c|}{}                                                & True Positive            & False Positive          & Accuracy                 & True Positive            & False Postive           & Accuracy                 \\ \hline
\multirow{2}{*}{Entire Dataset} & \multirow{2}{*}{ResNet-18}          & \multirow{2}{*}{96.52\%} & \multirow{2}{*}{0.60\%} & \multirow{2}{*}{97.19\%} & \multirow{2}{*}{99.86\%} & \multirow{2}{*}{0.08\%} & \multirow{2}{*}{99.88\%} \\
                                &                                     &                          &                         &                          &                          &                         &                          \\ \hline
\multirow{2}{*}{Manually Selected 200}        & ResNet-18                           & 96.00\%                  & 0.00\%                  & 98.00\%                  & 100.00\%                 & 0.00\%                  & 100.00\%                 \\ \cline{2-8} 
                                & \multicolumn{1}{l|}{Human Subjects} & 92.00\%                  & 8.00\%                  & 92.00\%                  & 98.00\%                  & 2.00\%                  & 98.00\%                  \\ \hline
\end{tabular}
\label{tab:results}
\end{table*}

A non-asymptomatic version of the iterated algorithm forms the basis of the definition of a confidence bound. The law therefore takes the form $\alpha(u) \propto \sqrt{log(log(u)n/\sigma)/n}$, where $u$ represents the integer that gives the number of the comparisons made and $\sigma \le 1$ is a fixed risk parameter \cite{kaufmann2016complexity}. The score is for the number of comparisons that are made and is registered together with the associated score's empirical permutation of $[n]$ in every round such that $\hat{\tau}_1 \geq \hat{\tau}_2 \dots \geq \hat{\tau}_n$. Then the following indices can be defined:
\begin{flalign}
    &d_1 = {argmin}_{i \in \{1,\dots,k-h\}} \hat{\tau}_i - \alpha_i, \\
    &d_2 = {argmax}_{i \in \{k+1+h,\dots,n\}} \hat{\tau}_i + \alpha_i
\end{flalign}

%PAUL COMMENT:  I think the last part of the above paragraph should be "...such that $\hat{\tau}_1 \geq \hat{\tau}_2 \dots \geq \hat{\tau}_n$. Then the following indices can be defined:"

Two additional indices, $b_1$ and $b_2$, defined as
\begin{flalign}
    &b_1 = {argmax}_{i \in \{d_1, (k-h+1), \dots, (k)\}}~ \alpha_i \\
    &b_2 = {argmax}_{i \in \{d_2, (k+1), \dots, (k+h)\}}~ \alpha_i
\end{flalign}
are the standard indices of the upper $(k-h)$ and the lower $(n-k-h)$ ranked images for the Lower-Upper Confidence Bound strategy represented in the work by Kaufman \textit{et al.} \cite{kalyanakrishnan2012pac}.

%PAUL COMMENT:  The sentence "Updates of indices $d_1$ and $d_2$ are done in each round of comparison for $h=0$ for each recovery of $k$ that is exact in this LUCB strategy..." doesn't make sense to me.  When do we compute $d_1$ and $d_2$ for $h=0$?  Also, I think it flows better of the equations for $b_1$ and $b_2$ are placed immediately following the sentence where they are introduced.

The main goal in this comparison is obtaining the two subsets $S_1$ and $S_2$ by making sufficient estimation of scores of the items. At each time instant, the algorithm determines which pair of items to present for comparison based on the outcomes of previous comparisons. The score's current estimations and the intervals of confidence associated with the scores are the parameters underlying the decision about which images to compare next in this strategy. As a result of comparing two items, Hamming-LUCB receives an independent draw of success in the view point of the comparator in response. The algorithm focuses on the upper ranked $k-h$ items  given as $\hat{S}_1=\{(1),…,(k-h)\}$ and the lower ranked $n-k-h$ items  given as $\hat{S}_2=\{(k+1+h),…,(n)\}$. Nevertheless, it does not disregard the rest of the items in between the two bounds within the sets. The confidence intervals of these items are kept below the confidence intervals for items $\hat{S}_1$ and $\hat{S}_2$.

%PAUL COMMENT:  For the paragraph below, the selection is based on the ranking and the confidence interval, not just the ranking, correct?

Pseudo-code for the Hamming-LUCB is shown in Algorithm \ref{alg:LUCB}. The algorithm terminates based on the associated stopping condition,
$\hat{\tau}_{d_1} - \alpha_{d_1} \geq \hat{\tau}_{d_2} + \alpha_{d_2}$. In our case, Hamming-LUCB can select the next two images from the dataset to compare based on the current rankings, thus converging to an ordering with many fewer comparisons than a brute force method.  

\begin{algorithm}[tbh!]              
  \caption{Hamming-LUCB}\label{alg:LUCB}
  \begin{algorithmic}
    \State \textit{n:} is the number of images\\
    $\hat{\tau}_{i}$ is the rank (score) of image $i$\\
    \textbf{$T_i$:} is the number of comparisons along with image $i$\\
    \textbf{$h$:} is the given tolerance for extracting the top-$k$ items defined by the Hamming distance\\
    \textbf{$\hat{S}_{1}$:} is the set of the $k - h$ top ranked images\\
    \textbf{$\hat{S}_{2}$:} is the set of the $n-k-h$ bottom ranked images
    \State \textbf{Initialization:} For every image $i \in [n]$, compare $i$ to an item $j$ chosen uniformly at random from $[n]$\textbackslash\{i\} and set $\hat{\tau}_{i}(1) = \mathbbm{1}$\{$i$ wins\} ($\mathbbm{1}$\{$i$ wins\} = 1 if $i$ is winner, $0$ otherwise), $T_i = 1$
    % \State \For {$i \in \{1,\ldots,n\}$}
    %             \State Create a permutation of [n] such that $\hat{\tau}_{1} \geq \dots \geq \hat{\tau}_{n}$
    %             \State $T_i = 1$ 
    %       \EndFor
           \State \While {\textsc{Termination Condition Not Satisfied}}
                \State Sort images, such that $\hat{\tau}_{1} \geq \dots \geq \hat{\tau}_{n}$
                \State Calculate $d_1$ and $d_2$
                \State Calculate $b_1$ and $b_2$
                \State \For{$j \in \{b_1, b_2\}$}
                           \State $T_j = T_j + 1 $
                           \State Compare $j$ to a random chosen image $k\in [n] $ \textbackslash \{j\}
                           \State update 
                           $\hat{\tau}_{j}=\frac{T_j - 1}{T_j}\hat{\tau}_{j} + \frac{1}{T_j} \mathbbm{1}$\{j wins\}    
                        \EndFor   
            \EndWhile
            \State Return $\hat{S}_{1}$ and $\hat{S}_{2}$  
  \end{algorithmic}
\end{algorithm} 

\subsubsection{Website Ratings Collection}
The inspiration of our website comes from that of the GIGGIF project for ranking emotions of GIFs\footnote{http://gifgif.media.mit.edu}. Figure \ref{fig::web} shows a screenshot of the website. 
The text ''Which of the following two faces looks MORE FAKE to you'' is displayed above two images. When the evaluator moves the mouse above either image, it is highlighted with a bounding box. The evaluator could choose to login using a registered account or stay as an anonymous evaluator. In this website, there are two instances of Hamming-LUCB running independently for two types of swapped faces. The probability of selecting either swapped type is 50\%. Over a three month period, we recruited volunteers to rate the images. When a new rater is introduced to the website, they first undergo a tutorial and example rating to ensure they understand the selection process. We collected 36,112 comparisons in total from more than  90 evaluators who created login accounts on the system. We note that anyone using the system anonymously (without logging in) was not tracked so it is impossible to know exactly how many evaluators used the website.

\section{Results}

To evaluate the performance of our classifier, we use five-fold cross validation to separate training and testing sets. We don't distinguish two types of swapped faces during training. In other words,  we mix the swapped faces generated using both methods during training, but we report prediction performance on each method separately. Table \ref{tab:results} gives the overall detection performance of our classifier for the entire dataset and for the 400 images that were ranked by human subjects. We also report the accuracy with which humans were able to select images as real or fake based on the pairwise ranking. That is, any fake images ranked in the top 50\% or any real images ranked in the bottom 50\% were considered as errors. From the table, we can see that both human subjects and the classifier achieve good accuracy when detecting swapped faces. Our classifier is able to achieve comparable results to human subjects in 200 manually selected representative images (100 fake, 100 real) for each method. 
% SUPERIOR? DID WE RUN A TEST FOR STATISTICAL DIFFERENCE? PERHAPS WE SHOULD USE A MCNEMAR TEST HERE?

\begin{figure*}[t]
\centering
\centerline{\includegraphics[width=1.0\linewidth]{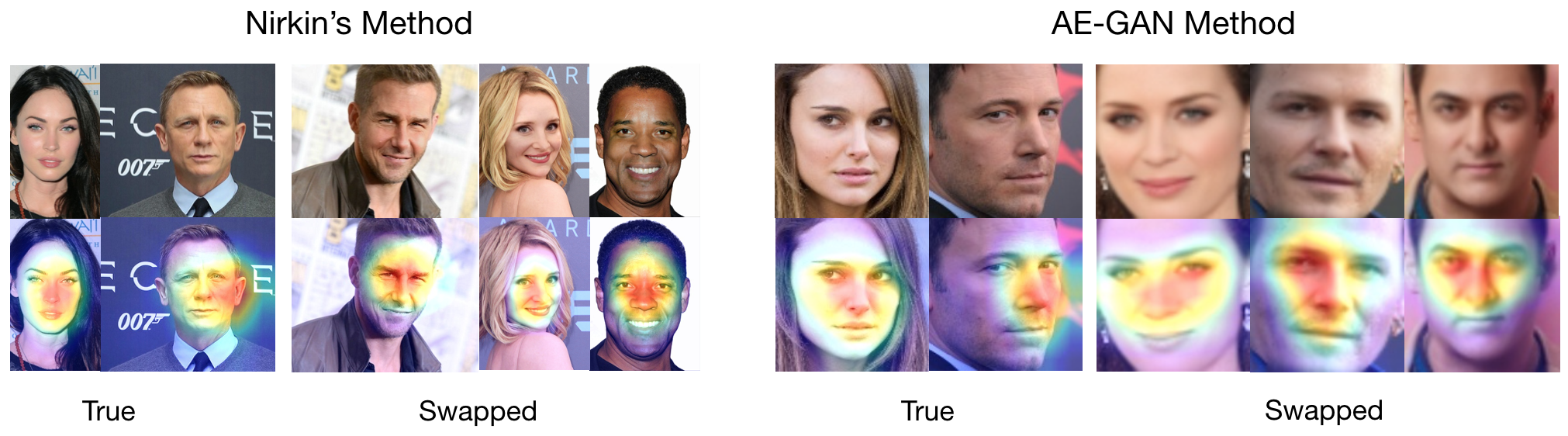}}
\caption{Grad CAM visualization of our proposed model on real and swapped faces. \textbf{Top Row:} original image. \textbf{Bottom Row:} original image with heatmap}
\label{fig::visual}
\end{figure*}

\subsection{Classification Accuracy}
As we mentioned above, we created two types of cropped images for each method. The AE-GAN method  contains minimal background and Nirkin's method contains more background. We can see from Table \ref{tab:results} that our classifier is able to detect face swapping better for the AE-GAN generated images---this holds true regardless of testing upon the entire dataset or using the manually selected 200. As we can see from Figure \ref{fig::dataset} (Right), the AE-GAN generates swapped faces that are slightly blurry, which we believe our model exploits for detection. On the other hand, Nirkin's method could generate swapped faces without a decrease in sharpness. Thus, it may require the model to learn more subtle features, such as looking for changes in lighting condition near the cropped face or stretching of facial landmarks to align the perspectives.

% Because the model can learn artifacts specific to each swapping technique, training with only one type of swapped faces may lead to an overly optimistic evaluation. In a real scenario, one would not know which kind of attack to expect. To investigate this scenario, we combined the images generated using both methods and train the model using all images. As we can see, the model is still robust with 98\% accuracy on the whole dataset. 

% Our model's performance on the top 200 image we picked is also robust and consistent with 100\% accuracy for the Auto-Encoder GAN and 99\% accuracy for Method-2. 

For version 1.0 of the dataset, we have collected more than 36,112 pairwise comparisons from more than 90 evaluators (approximately evenly split between each method). Human subjects may have different opinions about a pair of images, thus it requires many pairwise comparisons, especially for these images in the middle area. However, we can see human subjects still give a reasonable accuracy, especially for the AE-GAN method. It is interesting to see that both our classifier and human subjects perform better on the AE-GAN generated images.

\subsection{Classifier Visualization}

To elucidate what spatial area our classifiers are concentrating upon to detect an image as real or fake, we employ the Gradient-weighted Class Activation Mapping (Grad-CAM) visualization technique \cite{selvaraju2017grad}. This analysis helps mitigate the opaqueness of a neural network model and enhance explain-ability for applications in the domain of privacy and security. Grad-CAM starts by calculating the gradients of most dominant logit with respect to the last convolutional layer. The gradients are then pooled channel wise as weights. By inspecting these weighted activation channels, we can see which portions of the image have significant influence in classification.  For both types of generated swapped faces, our classifier focuses on the central facial area (\textit{i.e.},  the nose and eyes) rather than the background. This is also the case for real faces as we can see from Figure \ref{fig::visual}. We hypothesize that the classifier focuses on the nose and eyes because the most visible artifacts are typically contained here. It is interesting that the eyes and nose are focused upon by the classifier because human gaze also tends to focus on the eyes and nose when viewing faces \cite{guo2010human}.  
%We hypothesize that our classifier behaves similarly to human gaze patterns when doing manipulated faces forensic.

\subsection{Images ranking}

\begin{figure}[t]
\centerline{\includegraphics[width=1.0\linewidth]{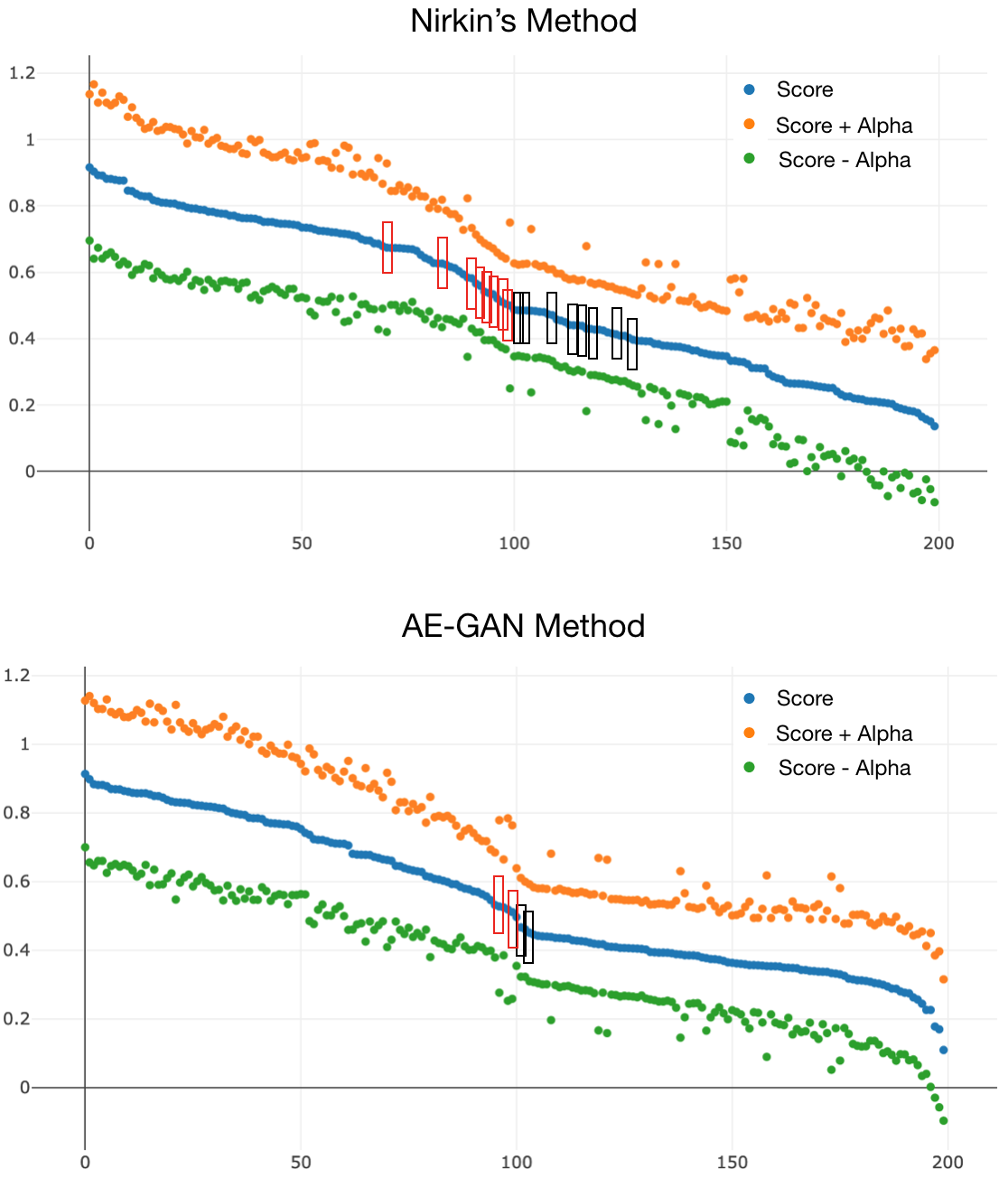}}
\caption{Human subjects rank of the manually selected 200 images. Left to right, from most real to most fake.}
\label{fig::ranking}
\end{figure}

Rather than reporting only accuracy of detecting swapped faces from human subjects, we also provide a ranking comparison. Ranking gives us more information to compare the models with, such as does the ResNet model similarly rate images that are difficult to rate for humans? Or, on the contrary, is the ranking from the model very different from human ratings?

Figure~\ref{fig::ranking} gives the overall ranking for faces generated using two methods using the Hamming-LUCB ranking from human evaluators. Red boxed points are false negatives, black boxed points are false positives. The alpha in the plot gives a confidence interval based on the Hamming-LUCB. As we can see, Human subjects have more difficulty classifying the faces generated using Nirkin's method. As mentioned, the AE-GAN generated faces are blurrier compared with Nirkin's method. Human subjects seemingly are able to learn such a pattern from previous experience. While some mistakes are present for the AE-GAN, these mistakes are very near the middle of the ranking. Swapped faces generated using Nirkin's method keep the original resolution and are more photo realistic---thus they are also more difficult to discern as fake. 

To compare the human ranking to our model, we need to process the outputs of the neural network. During training, the model learns a representation of the input data using convolutions. Instances belonging to different classes usually are pushed away in a high dimensional space. But this distance between two instances is not necessarily meaningful to interpret. Despite this, the output of the activation function can be interpreted as a relative probability that the instance belongs to each class. 
% May need to double check this part
% ERIC: I edited this to be a discussion of the max margin. 
%

\begin{figure}[t]
\centerline{\includegraphics[width=1.0\linewidth]{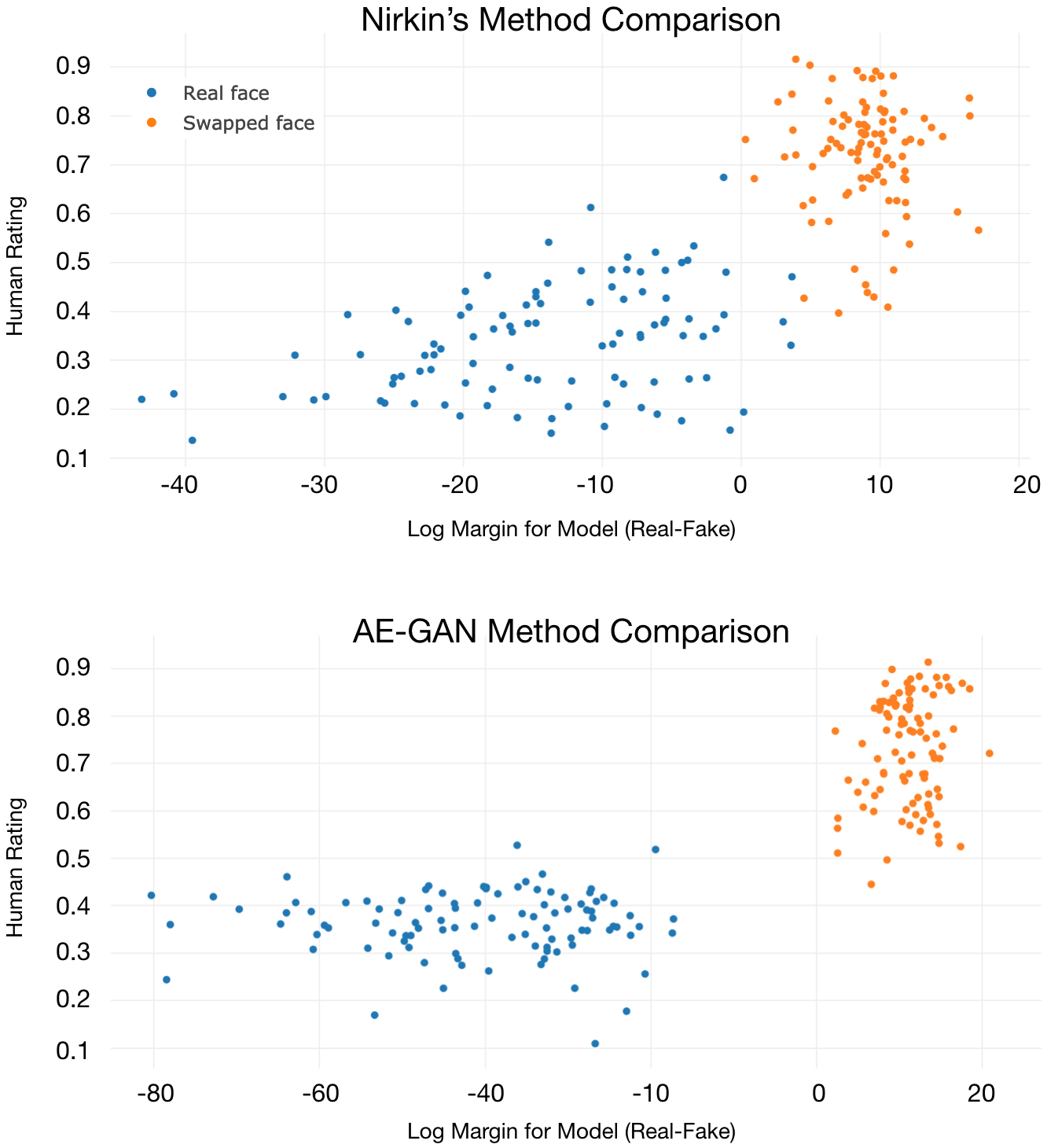}}
\caption{\textbf{Top:} Nirkin's Method linear correlation=0.7896, Spearman's rank order correlation=0.7579, p-value$<$0.01. \textbf{Bottom:} AE-GAN Method Linear correlation=0.8332, Spearman's rank order correlation=0.7576, p-value$<$0.01}
\label{fig::output_layer}
\end{figure}

%PAUL COMMENT:  This sentence "We assume for the odds ratio of the last full connected layer (before sent to the squashing function), the wider the margin, the more confident the classifier is that the instance is real or fake." needs editing.  The sentence is ambigous, and also doesn't seem to agree with how the ratio is being used (i.e., as a ranking predictor, not an uncertainty predictor).  Maybe change the last part to "with a wider margin (higher value) indicates the classifier is more confident in its assessment that the instance is real or fake, and hence can be used as a proxy for ranking the images."
We assume for the odds ratio of the last full connected layer (before sent to the squashing function), the wider the margin, the more confident the classifier is that the instance is real or fake. Fig. \ref{fig::output_layer} gives the comparison of log margin of our model and human rating for the 200 faces.  For Nirkin's Method, the linear correlation is 0.7896 and Spearman's rank order correlation is 0.7579. For the AE-GAN Method, the linear correlation 0.8332 and Spearman's rank order correlation is 0.7576. This indicates that the uncertainty level of our model and human subjects is consistent, though not perfect. This consistency is encouraging because it shows the model learns not only a binary threshold, but captures similarity in ranking of the images from most fake to most real. We anticipate that future work can further improve upon this ranking similarity.

%% I need to double check this results!

\section{Conclusion}

In this study, we investigated using deep transfer learning for swapped face detection. For this purpose, we created the largest public face swapping detection data set using still images. Moreover, the data set has around 1,000 real images for each individual (publicly known largest), which is beneficial for models like the AE-GAN face swapping method. We use this data set to inform the design and evaluation of a classifier and the results show the effectiveness of the model for detecting swapped faces. More importantly, we compare the performance of our model with human subjects. We designed and deployed a website to collect pairwise comparisons for 400 carefully picked images from our data set. Approximate ranking is calculated based on these comparisons. We compared the ranking of our deep learning model and find that it shows good correspondence to human ranking. We hope this work will assist in the creation and evaluation of future image forensics algorithms. 

%PAUL COMMENT:  "Moreover, the dataset has around 1,000 real images for each individual (publicly known largest)..."  Should this be "...each individual (largest publicly known)..."?  I'm not sure what the statement in parenthesis is trying to indicate since it's already stated that the created data set is the largest public face swapping detection dataset.

\bibliographystyle{abbrv}
\bibliography{sfd}

\begin{thebibliography}{10}

\bibitem{Deepfake15:online}
Deepfakes porn has serious consequences - bbc news.
\newblock \url{https://www.bbc.com/news/technology-42912529}.
\newblock (Accessed on 05/28/2019).

\bibitem{GitHubsh94:online}
Github - shaoanlu/faceswap-gan: A denoising autoencoder + adversarial losses
  and attention mechanisms for face swapping.
\newblock \url{https://github.com/shaoanlu/faceswap-GAN}.
\newblock (Accessed on 05/11/2019).

\bibitem{agarwal2017swapped}
A.~Agarwal, R.~Singh, M.~Vatsa, and A.~Noore.
\newblock Swapped! digital face presentation attack detection via weighted
  local magnitude pattern.
\newblock In {\em 2017 IEEE International Joint Conference on Biometrics
  (IJCB)}, pages 659--665. IEEE, 2017.

\bibitem{bay2006surf}
H.~Bay, T.~Tuytelaars, and L.~Van~Gool.
\newblock Surf: Speeded up robust features.
\newblock In {\em European conference on computer vision}, pages 404--417.
  Springer, 2006.

\bibitem{bitouk2008face}
D.~Bitouk, N.~Kumar, S.~Dhillon, P.~Belhumeur, and S.~K. Nayar.
\newblock Face swapping: automatically replacing faces in photographs.
\newblock In {\em ACM Transactions on Graphics (TOG)}, volume~27, page~39. ACM,
  2008.

\bibitem{blanz2004exchanging}
V.~Blanz, K.~Scherbaum, T.~Vetter, and H.-P. Seidel.
\newblock Exchanging faces in images.
\newblock In {\em Computer Graphics Forum}, volume~23, pages 669--676. Wiley
  Online Library, 2004.

\bibitem{chen2019face}
D.~Chen, Q.~Chen, J.~Wu, X.~Yu, and T.~Jia.
\newblock Face swapping: Realistic image synthesis based on facial landmarks
  alignment.
\newblock {\em Mathematical Problems in Engineering}, 2019, 2019.

\bibitem{dale2011video}
K.~Dale, K.~Sunkavalli, M.~K. Johnson, D.~Vlasic, W.~Matusik, and H.~Pfister.
\newblock Video face replacement.
\newblock In {\em ACM Transactions on Graphics (TOG)}, volume~30, page 130.
  ACM, 2011.

\bibitem{deng2009imagenet}
J.~Deng, W.~Dong, R.~Socher, L.-J. Li, K.~Li, and L.~Fei-Fei.
\newblock Imagenet: A large-scale hierarchical image database.
\newblock In {\em 2009 IEEE conference on computer vision and pattern
  recognition}, pages 248--255. Ieee, 2009.

\bibitem{goodfellow2014generative}
I.~Goodfellow, J.~Pouget-Abadie, M.~Mirza, B.~Xu, D.~Warde-Farley, S.~Ozair,
  A.~Courville, and Y.~Bengio.
\newblock Generative adversarial nets.
\newblock In {\em Advances in neural information processing systems}, pages
  2672--2680, 2014.

\bibitem{guo2010human}
K.~Guo, D.~Tunnicliffe, and H.~Roebuck.
\newblock Human spontaneous gaze patterns in viewing of faces of different
  species.
\newblock {\em Perception}, 39(4):533--542, 2010.

\bibitem{heckel2018approximate}
R.~Heckel, M.~Simchowitz, K.~Ramchandran, and M.~J. Wainwright.
\newblock Approximate ranking from pairwise comparisons.
\newblock {\em arXiv preprint arXiv:1801.01253}, 2018.

\bibitem{LFWTech}
G.~B. Huang, M.~Ramesh, T.~Berg, and E.~Learned-Miller.
\newblock Labeled faces in the wild: A database for studying face recognition
  in unconstrained environments.
\newblock Technical Report 07-49, University of Massachusetts, Amherst, October
  2007.

\bibitem{kalyanakrishnan2012pac}
S.~Kalyanakrishnan, A.~Tewari, P.~Auer, and P.~Stone.
\newblock Pac subset selection in stochastic multi-armed bandits.
\newblock In {\em ICML}, volume~12, pages 655--662, 2012.

\bibitem{kaufmann2016complexity}
E.~Kaufmann, O.~Capp{\'e}, and A.~Garivier.
\newblock On the complexity of best-arm identification in multi-armed bandit
  models.
\newblock {\em The Journal of Machine Learning Research}, 17(1):1--42, 2016.

\bibitem{khodabakhsh2018fake}
A.~Khodabakhsh, R.~Ramachandra, K.~Raja, P.~Wasnik, and C.~Busch.
\newblock Fake face detection methods: Can they be generalized?
\newblock In {\em 2018 International Conference of the Biometrics Special
  Interest Group (BIOSIG)}, pages 1--6. IEEE, 2018.

\bibitem{korshunov2018deepfakes}
P.~Korshunov and S.~Marcel.
\newblock Deepfakes: a new threat to face recognition? assessment and
  detection.
\newblock {\em arXiv preprint arXiv:1812.08685}, 2018.

\bibitem{korshunova2017fast}
I.~Korshunova, W.~Shi, J.~Dambre, and L.~Theis.
\newblock Fast face-swap using convolutional neural networks.
\newblock In {\em Proceedings of the IEEE International Conference on Computer
  Vision}, pages 3677--3685, 2017.

\bibitem{mahajan2017swapitup}
S.~Mahajan, L.-J. Chen, and T.-C. Tsai.
\newblock Swapitup: A face swap application for privacy protection.
\newblock In {\em 2017 IEEE 31st International Conference on Advanced
  Information Networking and Applications (AINA)}, pages 46--50. IEEE, 2017.

\bibitem{natsume2018rsgan}
R.~Natsume, T.~Yatagawa, and S.~Morishima.
\newblock Rsgan: face swapping and editing using face and hair representation
  in latent spaces.
\newblock {\em arXiv preprint arXiv:1804.03447}, 2018.

\bibitem{nirkin2018face}
Y.~Nirkin, I.~Masi, A.~T. Tuan, T.~Hassner, and G.~Medioni.
\newblock On face segmentation, face swapping, and face perception.
\newblock In {\em 2018 13th IEEE International Conference on Automatic Face \&
  Gesture Recognition (FG 2018)}, pages 98--105. IEEE, 2018.

\bibitem{perez2003poisson}
P.~P{\'e}rez, M.~Gangnet, and A.~Blake.
\newblock Poisson image editing.
\newblock {\em ACM Transactions on graphics (TOG)}, 22(3):313--318, 2003.

\bibitem{rossler2019faceforensics++}
A.~R{\"o}ssler, D.~Cozzolino, L.~Verdoliva, C.~Riess, J.~Thies, and
  M.~Nie{\ss}ner.
\newblock Faceforensics++: Learning to detect manipulated facial images.
\newblock {\em arXiv preprint arXiv:1901.08971}, 2019.

\bibitem{schmidhuber2015deep}
J.~Schmidhuber.
\newblock Deep learning in neural networks: An overview.
\newblock {\em Neural networks}, 61:85--117, 2015.

\bibitem{selvaraju2017grad}
R.~R. Selvaraju, M.~Cogswell, A.~Das, R.~Vedantam, D.~Parikh, and D.~Batra.
\newblock Grad-cam: Visual explanations from deep networks via gradient-based
  localization.
\newblock In {\em Proceedings of the IEEE International Conference on Computer
  Vision}, pages 618--626, 2017.

\bibitem{zhang2017automated}
Y.~Zhang, L.~Zheng, and V.~L. Thing.
\newblock Automated face swapping and its detection.
\newblock In {\em 2017 IEEE 2nd International Conference on Signal and Image
  Processing (ICSIP)}, pages 15--19. IEEE, 2017.

\end{thebibliography}

\end{document}